# Machine Learning Distinguishes Neurosurgical Skill Levels in a Virtual Reality Tumor Resection Task

**Samaneh Siyar, MSc,[1,2] Hamed Azarnoush, PhD,[1,2] Saeid Rashidi, PhD,[3] Alexandre Winkler-Schwartz, MD,[1] Vincent Bissonnette, MD,[1] Nirros Ponnudurai, BEng,[1] Rolando F. Del Maestro, MD, PhD[1]**

1. Neurosurgical Simulation Research and Training Centre, Department of Neurosurgery, Montreal Neurological Institute and Hospital, McGill University, Canada

2. Department of Biomedical Engineering, Amirkabir University of Technology (Tehran Polytechnic), Iran

3. Science and Research Branch, Islamic Azad University, Iran

**Details of previous presentations:**

A portion of this work has been presented in the form of an abstract at the Computer Assisted Radiology and Surgery congress, June 19-23, 2018, Berlin.

**Disclosure of financial support:**

This work was supported by the Di Giovanni Foundation, the Montreal English School Board, the B-Strong Foundation, the Colannini Foundation, the Montreal Neurological Institute and Hospital and the McGill Department of Orthopedics. Samaneh Siyar is a Visiting Scholar in the Neurosurgical Simulation Research and Training Centre. Dr. H. Azarnoush previously held the Postdoctoral Neuro-Oncology Fellowship from the Montreal Neurological Institute and Hospital and is a Visiting Professor in the Neurosurgical Simulation Research and Training Centre. Dr. Winkler-Schwartz holds a Robert Maudsley Fellowship from the Royal College of Physicians and Surgeons of Canada and Nirros Ponnudurai is supported by a Heffez Family Bursary. Dr. Del Maestro is the William Feindel Emeritus Professor in Neuro-Oncology at McGill University.

**Acknowledgments**

We thank all the medical students, residents, and neurosurgeons from the Montreal Neurological Institute and Hospital and at other institutions who participated in this study. We would also like to thank Robert DiRaddo, Group Leader, Simulation, Life Sciences Division, National Research Council Canada at Boucherville and his team, including Denis Laroche, Valérie Pazos, Nusrat Choudhury and Linda Pecora for their support in the development of the scenarios used in these studies and all the members of the Simulation, Life Sciences Division, National Research Council Canada.

**Corresponding author:**

Dr. Hamed Azarnoush, PhD

Visiting professor, Neurosurgical Simulation Research and Training Centre

Montreal Neurological Hospital, McGill University

3801 University, E2.89

Montreal, Quebec, Canada H3A 2B4

Tel: 514-934-1934 ext. 36733

Email: hamed.azarnoush@gmail.com

# Abstract:

**Background:** Virtual reality simulators and machine learning have the potential to augment understanding, assessment and training of psychomotor performance in neurosurgery residents.

**Objective:** This study outlines the first application of machine learning to distinguish "skilled" and "novice" psychomotor performance during a virtual reality neurosurgical task.

**Methods:** Twenty-three neurosurgeons and senior neurosurgery residents comprising the "skilled" group and 92 junior neurosurgery residents and medical students the "novice" group. The task involved removing a series of virtual brain tumors without causing injury to surrounding tissue. Over 100 features were extracted and 68 selected using t-test analysis. These features were provided to 4 classifiers: K-Nearest Neighbors, Parzen Window, Support Vector Machine, and Fuzzy K-Nearest Neighbors. Equal Error Rate was used to assess classifier performance.

**Results:** Ratios of train set size to test set size from 10% to 90% and 5 to 30 features, chosen by the forward feature selection algorithm, were employed. A working point of 50% train to test set size ratio and 15 features resulted in an equal error rates as low as 8.3% using the Fuzzy K-Nearest Neighbors classifier.

**Conclusion:** Machine learning may be one component helping realign the traditional apprenticeship educational paradigm to a more objective model based on proven performance standards.

# 1. Introduction

Virtual reality simulators have been proposed as tools to understand, assess and train neurosurgery residents.[1-5] An important element of simulator performance is the capacity of simulators to distinguish operator expertise. Most studies on operator performance have utilized "metrics."[6-16] A useful definition of "metrics" is that they are standards of reference by which performance, efficiency, and progress can be assessed. Individual metric can be used to assess aspects of operator performance. Applied forces,[9,17-21] bimanual dexterity,[15,22,23] and stress[23] have all been studied. An operator's performance metric(s) can be compared with previously defined proficiency benchmarks and that operator is placed into 1 of 2 or more groups with specific levels of psychomotor expertise.[24,25] Neurosurgical tasks are complicated, involving multiple cognitive processes and psychomotor skills, and larger sets of more complex and interacting metrics may be required to differentiate groups.

Artificial intelligence utilizing machine learning algorithms (classifiers) have the capacity to use extensive data sets involving numbers of features to separate groups.[26-30] Machine learning has been reviewed in neurosurgery[28] and to characterize performance during otolaryngology and dental virtual reality procedures.[31-36] Machine learning classifiers have not been utilized to differentiate "skilled" and "novice" neurosurgical psychomotor performance using a virtual reality simulator with haptic feedback. The question addressed in this communication is "do the 4 classifiers utilized in our study, K-Nearest Neighbors, Parzen Window, Support Vector Machine, and Fuzzy K-Nearest Neighbors have the ability to differentiate "skilled" from "novice" neurosurgical psychomotor performance using a virtual reality simulation platform?"

# 2. Methods

## 2.1. Subjects

115 individuals including 16 board certified practicing neurosurgeons from 3 institutions and 7 senior residents (PGY 4-6) from one university made up the expert group (n=23). Eight junior residents (PGY 1-3) and 84 medical students comprised the novice group (n=92). No participant had had previous experience with the simulator utilized and participants signed an approved Research Ethics Board consent.

## 2.2. NeuroVR

The NeuroTouch, now known as NeuroVR (CAE Healthcare, Montreal, Canada), virtual reality simulation platform was used.[5] Tumor resections were performed using the simulated ultrasonic aspirator held in the dominant hand (Fig. 1A).

## 2.3. Simulation Scenarios

Figure 1B outlines the 6 scenarios used in this study utilized data involving the resection of 9 identical simulated brain tumors on 2 occasions (total of 18 procedures) separated by removal of tumors with different complexities. The simulated operative procedure utilized for these studies can be seen in electronic Supplementary Material 1 in a previous publication.[9] To maximize tumor differences and increase participant difficulty, each of the 6 scenarios utilized had 3 tumors of varying complexities involving color (black, glioma-like and white: similar to background) and Young's modulus stiffness (3 kPa, soft, 9 kPa, medium and 15 kPa, hard). The background with soft tumor stiffness, 3 kPa represented the surrounding 'normal' white matter (Fig. 1C). Scenario 1 included 3 black tumors with different stiffness. Scenario 2 included 3 glioma-like tumors with different stiffness and Scenario 3 included white tumors with different stiffness. In Scenarios 4 through 6, all three simulated tumors included in each scenario had the same stiffness but were visually different. Scenario 4 included 3 soft tumors with different visual properties. Scenario 5 included 3 medium stiffness tumors with different visual properties and Scenario 6 included 3 hard tumors with different visual properties.[9] Three minutes was allowed for each tumors removal with a 1-minute rest time given between tumor resections to decrease fatigue. The trial involved 54 minutes of active tumor resection, 71 minutes in total. To develop procedure familiarity operators resected a practice scenario but this data was not used. Participants were unaware of study purpose or metrics utilized and were instructed to resect each tumor with minimal removal of the background tissue.

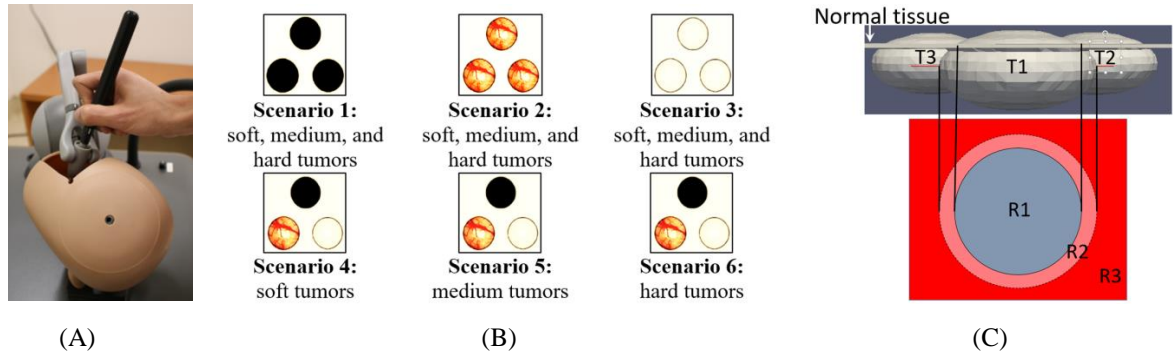

FIGURE 1. The hand position of the operator holding the simulated ultrasonic aspirator (A), the 6 simulated tumor scenarios with tumor color and sequence (B) and lateral view of the brain tumor geometry and ellipsoidal shape utilized in each scenario demonstrating the three identical tumors, tumor buried underneath simulated "normal" tissue and the R1 and the R2 plus R3 regions studied (C).

## 2.4. Feature extraction

The 3 processing steps, including feature extraction, feature selection, and classification are seen in Figure 2. Features may be considered inputs which are provided to machine learning algorithms to help define level of expertise. The simulator recorded signals including tool tip coordinates, tool tip orientation angles, contact force between virtual tool and virtual tissue and foot pedal activation state versus time. Although these signals provide useful information, previous investigations on developing a model for psychomotor performance for virtual reality tumor resections have outlined complex interacting human and task factors involved in differentiating skilled and novice performance.[20] Different parametric features could be extracted from these and other derived signal features with the goal to differentiate the skilled and novice groups.[37]

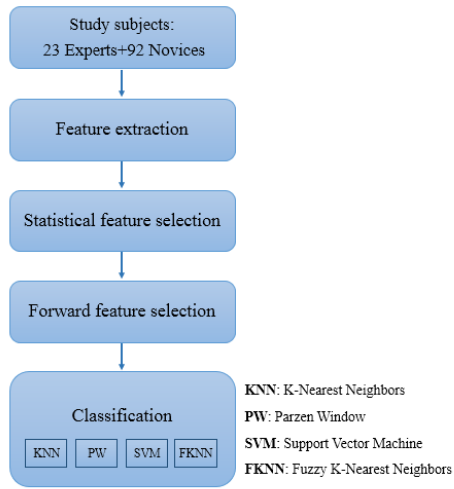

FIGURE 2. Flowchart of the proposed feature selection and classification.

## 2.4.1. Motion-based features

To obtain motion-based parametric features, first, signals such as velocity (first derivative), acceleration (second derivative) and jerk (third derivative) for position and angle signals were acquired. Then, based on signal features, parametric temporal, and spatial features were extracted.

Speed

Velocity was computed as the first derivative of motion profile and then speed was considered as the magnitude of the velocity profile. Some features based on the speed values used included mean speed, maximum speed, number of local maximum in the velocity vector and movement arrest period ratio.[38]

Acceleration

Acceleration was computed as the second derivative of the motion profile. Features based on the acceleration signal measured included mean acceleration, maximum acceleration and the integral of the acceleration vector ($IAV$)[39] as given by:

$$IAV = \int_0^T \sqrt{\left(\frac{d^2x}{dt^2}\right)^2 + \left(\frac{d^2y}{dt^2}\right)^2 + \left(\frac{d^2z}{dt^2}\right)^2} \; dt \tag{1}$$

where $x$, $y$ and $z$ are Cartesian coordinates and $T$ is the duration of the task.

Jerk

Jerk is defined as the third derivative of motion profile applied for motor skill assessment. A normalized three dimensional jerk[39] metric is used in this study, given by:

$$Jerk_{norm} = \sqrt{\frac{T^5}{2A_m^2} \int_0^T \left(\frac{d^3x}{dt^3}\right)^2 + \left(\frac{d^3y}{dt^3}\right)^2 + \left(\frac{d^3z}{dt^3}\right)^2} \, dt \qquad (2)$$

where T is task completion time and $A_m$ is the amplitude of the motion.

## 2.4.2. Force-based features

Not being able to measure forces applied by the neurosurgical aspirator during patient related procedures has limited our understanding of the forces that the human brain is exposed to by this instrument. The simulation platform utilized has the ability to analyze force feedback generated by the haptic device. This data has been utilized to create force pyramids and force heat maps to assess psychomotor function, automaticity, and force fingerprints for virtual reality tumor resections.[9,17,18,20] Force-based features extracted in this study comprise force derivatives, integral of the force, the range of the applied forces and the interquartile range, i.e., the first quartile subtracted from the third quartile. In addition to the above mentioned force-based features, parametric features including temporal and spatial features were also extracted from the force signal and its derivatives. We also used the 2 features proposed previously to indicate consistency,[40] given by:

$$df_{metric} = \sqrt{\frac{T}{2f_{iqr}^2} \int_0^T \left(\frac{df}{dt}\right)^2 dt} \qquad (3)$$

$$d^2f_{metric} = \sqrt{\frac{T^3}{2f_{iqr}^2} \int_0^T \left(\frac{d^2f}{dt^2}\right)^2 dt} \qquad (4)$$

and one feature to indicate the smoothness of the force application,[40] given by:

$$d^3f_{metric} = \sqrt{\frac{T^5}{2f_{iqr}^2} \int_0^T \left(\frac{d^3f}{dt^3}\right)^2 dt} \qquad (5)$$

where $T$ is task completion time and $f_{iqr}$ is the interquartile range of the force profile.

We started with a list of over 100 parametric features many of which were eliminated in the subsequent feature selection process. The list of all signal features is included in Table 1.

TABLE 1. List of signal features

| | |
|---|---|
| $x(t)$: position in the $x$-direction | $j_z(t) = \frac{da_z(t)}{dt}$ : jerk in the $z$-direction |
| $y(t)$: position in the $y$-direction | $j_f(t) = \frac{da_f(t)}{dt}$ : third derivative of force signal |
| $z(t)$: position in the $z$-direction | $Roll(t)$ : Rotation around the front-to-back axis |
| $f(t)$: $force$ | $v_{Roll}(t) = \frac{dRoll(t)}{dt}$ : first derivative of Roll signal |
| $v_x(t) = \frac{dx(t)}{dt}$ : velocity in the $x$-direction | $a_{Roll}(t) = \frac{dv_{Roll}(t)}{dt}$ : second derivative of Roll signal |
| $v_y(t) = \frac{dy(t)}{dt}$ : velocity in the $y$-direction | $j_{Roll}(t) = \frac{da_{Roll}(t)}{dt}$ : third derivative of Roll signal |
| $v_z(t) = \frac{dz(t)}{dt}$ : velocity in the $z$-direction | $Pitch(t)$: Rotation around the side-to-side axis |
| $v_f(t) = \frac{df(t)}{dt}$ : first derivative of the force signal | $v_{Pitch}(t) = \frac{dPitch(t)}{dt}$ : first derivative of Pitch signal |
| $V(t) = \sqrt{\frac{dx^2}{dt} + \frac{dy^2}{dt} + \frac{dz^2}{dt}}$ : magnitude of velocity | $a_{Pitch}(t) = \frac{dv_{Pitch}(t)}{dt}$ : second derivative of Pitch signal |
| $a_x(t) = \frac{dv_x(t)}{dt}$ : acceleration in the $x$-direction | $j_{Pitch}(t) = \frac{da_{pitch}(t)}{dt}$ : third derivative of Pitch signal |
| $a_y(t) = \frac{dv_y(t)}{dt}$ : acceleration in the $y$-direction | $Yaw(t)$: Rotation around the vertical axis |
| $a_z(t) = \frac{dv_z(t)}{dt}$ : acceleration in the $z$-direction | $v_{Yaw}(t) = \frac{dYaw(t)}{dt}$ : first derivative of Yaw signal |
| $a_f(t) = \frac{dv_f(t)}{dt}$ : second derivative of force signal | $a_{Yaw}(t) = \frac{dv_{Yaw}(t)}{dt}$ : second derivative of Yaw signal |
| $j_x(t) = \frac{da_x(t)}{dt}$ : jerk in the $x$-direction | $j_{yaw}(t) = \frac{da_{Yaw}(t)}{dt}$ : third derivative of Yaw signal |
| $j_y(t) = \frac{da_y(t)}{dt}$ : jerk in the $y$-direction | |

## 2.5. Feature normalization

Since the parametric feature values are not in the same order of size for comparison and to train classifiers, the obtained features were normalized exponentially:

$$Z_i = e^{-\frac{x_i}{max(x)}}$$

(6)

where $Z_i$ is the normalized value and $x_i$ is a data point ($x_1, x_2, \ldots, x_n$).

## 2.6. Feature selection

Feature selection follows feature extraction to decrease computational complexity and maintain classifier performance.[41] Irrelevant features are identified and only useful ones are provided to classifiers since irrelevant features may result in overfitting and increase resource use. In these techniques, an efficient search strategy is adopted to determine a feature subset. Then the new selected subset can be evaluated based on evaluation criteria.[42] Feature selection was carried out in 2 steps, first, the features with a defined statistical differentiation were identified and second, features improving classifier performance were selected.

### 2.6.1. Statistical feature selection

For each feature a t-test was applied and the resultant p-values were compared for all features as a measure of the usefulness of each individual feature to separate groups. Among the extracted preliminary features, 68 were able to differentiate the 2 groups with a statistically significant difference of $p < 0.05$ provided in Table 2.

TABLE 2. List of 68 selected parametric features that provide the best classification. The best 30 features are marked by one asterisk (*) and the best 15 features by two asterisks (**).

| | | | |
|---|---|---|---|
| 1 | $\sum t(j_x \le 0)/T$     $T$: task completion time | *35 | $N_{extermum}(f)$ |
| 2 | $\sum N(f > 0.1)$ | *36 | $\sum_{R_4} f$ <br> $R_4$: region beneath the tumor bulk |
| *3 | $std(f)/std(v_x)$     $std$: standard deviation | 37 | $\max(f) - \min(f)$ |
| 4 | $(\max(v_x) - \min(v_x)) * (\max(v_y) - \min(v_y)) * (\max(v_z) - \min(v_z))$ | 38 | $std(f)$     $std$: standard deviation |
| 5 | $iqr(f)$ | 39 | $iqr(x) * iqr(y) * iqr(z)$ |
| 6 | $\sqrt{\dfrac{T^3}{2(iqr(f))^2}\sum a_f{}^2}$ | **40 | $\sqrt{\dfrac{T}{2(iqr(f))^2}\sum v_f{}^2}$ |
| 7 | $(\sum_{i=1}^{N-1}|f_{i+1} - f_i|)/T$ | 41 | $\sum_i \sqrt{a_{xi}^2 + a_{yi}^2 + a_{zi}^2}$ |
| **8 | $(\sum_{i=1}^{N-1}|v_{i+1} - v_i|)/T(std(f))$ | *42 | $(\sum_{i=1}^{N-1}|a_{i+1} - a_i|)/T$ |
| **9 | $(\sum t(\max(a_x)))/T$ | 43 | $\int_{t_1}^{t_2}|f|$ <br> $t_1$: start point of signal peak   $t_2$: end point of signal peak |
| *10 | $N_{zero-cross}(v_x)$ | *44 | $(\sum t(\min(a_x)))/T$ |
| *11 | $(\sum t(\min(a_y)))/T$ | **45 | $(\sum t(\max(a_y)))/T$ |
| 12 | $(\sum t(\max(a_z)))/T$ | 46 | $N_{zero-cross}(v_y)$ |
| 13 | $N_{extermum}(Pitch)$ | **47 | $(\sum t(\min(a_z)))/T$ |
| *14 | $(\sum t(\min(a_f)))/T$ | *48 | $(\sum t(\max(a_f)))/T$ |
| 15 | $mean(V)$ | 49 | $\max(V)$ |
| *16 | $std(f)/std(v_z)$     $std$: standard deviation | **50 | $\sqrt{\dfrac{T^5}{2(iqr(f))^2}\sum j_f{}^2}$ |
| **17 | $\sum f_{low\ frequency} / \sum f_{high\ frequency}$ | **51 | $std(f)/std(v_y)$   $std$: standard deviation |
| 18 | $(\sum t(\max(z)))/T$ | *52 | $N_{minimum}(x)$ |
| 19 | $N_{extermum}(v_f)$   $N_{extermum}$: Number of extermum points | **53 | $N_{minimum}(v_x)$ |
| **20 | $N_{minimum}(a_x)$ | **54 | $(\sum t(\max(f)))/\sum t(\min(f))$ |
| 21 | $N_{minimum}(a_y)$ | 55 | $N_{extermum}(a_f)$ |
| 22 | $N_{max}(x) + N_{max}(y) + N_{max}(z)$ | 56 | $\sum t(v_f \geq 0) / \sum t(v_f) \leq 0$ |
| **23 | $N_{extermum}(x)$ | 57 | $N_{min}(x) + N_{min}(y) + N_{min}(z)$ |
| 24 | $N_{extermum}(z)$ | 58 | $N_{extermum}(y)$ |
| 25 | $(\sum_{i=1}^{N-1} |Pitch_{i+1} - Pitch_i|)/T$ | *59 | $\sum_{i=1}^{N-1} |Yaw_{i+1} - Yaw_i|$ |
| 26 | $(\sum_{i=1}^{N-1} |v_{Roll_{i+1}} - v_{Roll_i}|)/T$ | 60 | $(\sum_{i=1}^{N-1} |v_{Pitch_{i+1}} - v_{Pitch_i}|)/T$ |
| *27 | $mean(Roll) * T/\max(v_{Roll}) - \min(v_{Roll})$ | 61 | $mean(Pitch) * T/\max(v_{Pitch}) - \min(v_{Pitch})$ |
| 28 | $N_{min}(Yaw) + N_{min}(Pitch) + N_{min}(Roll)$ | **62 | $N_{max}(Yaw) + N_{max}(Pitch) + N_{max}(Roll)$ |
| 29 | $N_{extermum}(Pitch)$ | 63 | $N_{extermum}(Yaw)$ |
| 30 | $N_{extermum}(v_{Yaw})$ | **64 | $N_{extermum}(Roll)$ |
| **31 | $N_{extermum}(v_{Roll})$ | *65 | $N_{extermum}(v_{Pitch})$ |
| *32 | $\sum t(\max(Pitch)) - \sum t(\min(Pitch)/T$ | 66 | $\sum_{i=1}^{N-1} |j_{Pitch_{i+1}} - j_{Pitch_i}|/mean|j_{Pitch}|$ |
| 33 | *Frequency of pedal activation* | 67 | $\sum t(\max(Yaw)) - \sum t(\min(Yaw)/T$ |
| 34 | $\sum_{R3} f$ | 68 | $\sum_{R1} f$ |

### 2.6.2. Forward feature selection

To find the most relevant features, forward feature selection, backward feature selection and genetic algorithms were applied. The forward feature selection algorithm provided best results and was therefore employed. This algorithm starts with an empty set and adds features one by one outlining the best feature set of particular size.[42] This algorithm was applied to rank the best 5, 10, 15, 20, 25 and 30 features of 68 selected utilizing the statistics previously employed. The feature selection was done irrespective of the subsequent classifiers to be used in the next stage.

## 2.7. Classification

Four classifiers, K-Nearest Neighbors with k = 7, Parzen Window, Support Vector Machine and Fuzzy K-Nearest Neighbors with k=7 were applied to classify skilled and novice groups. To improve the configuration of the train and test sets, train set was selected randomly and increasingly from 10% to 90% of all data. For cross validation of each train set size, classification process was repeated 20 times, each time with a randomly selected train set of that size. Equal error rate (EER) is obtained when sensitivity and specificity become equal. A classifier may have a good sensitivity and a poor specificity and vice versa. Equal error rate is a commonly-used measure to evaluate classifier performance since it evaluates its sensitivity and specificity at the same time.[43] Equal error rates were measured for different working points of both different train set sizes and different number of premier features.

# 3. Results

## 3.1. Influence of train set size

All classifiers were applied on 5, 10, 15, 20, 25 and 30 superior features for 20 iterations. Figure 3 outlines the ability of each classifier to discriminate between skilled and novice groups in the 6 different scenarios for different train set sizes between 10% and 90% with 10% increments. The overall trend of classifiers' performance are similar for all scenarios. We used 50% train set size as the working point, since increasing the train set size did not significantly improve classifier performance. For this train set size, the minimum equal error rate value in Scenario 1, 2, 4, and 5 were obtained for the Fuzzy K-Nearest Neighbors classifier ($EER_1$=9.6%, $EER_2$=13.7%, $EER_4$=11.1% $EER_5$=12.0%). In scenario 3 the Parzen Window classifier resulted in minimum equal error rate ($EER_3$=10.8%) and in scenario 6, the minimum equal error rate was acquired for K-Nearest Neighbors classifier ($EER_6$=14.3%).

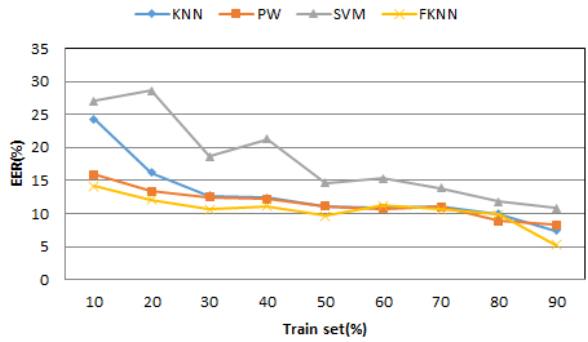

(A)

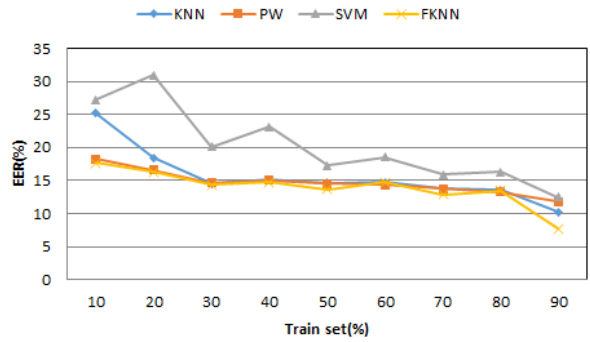

(B)

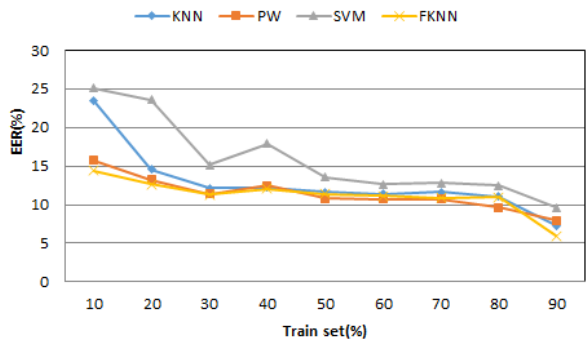

(C)

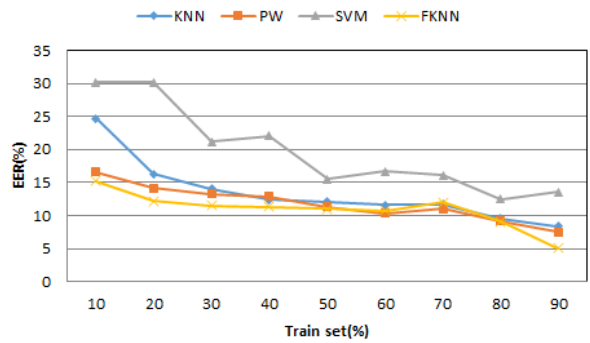

(D)

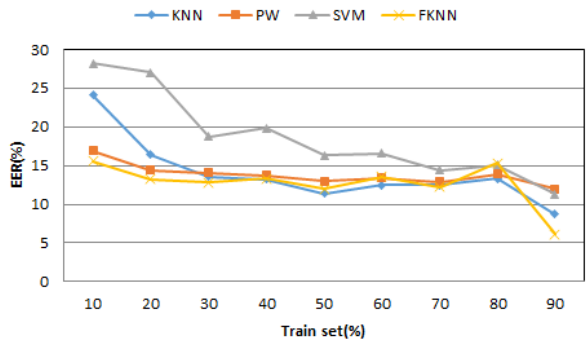

(E)

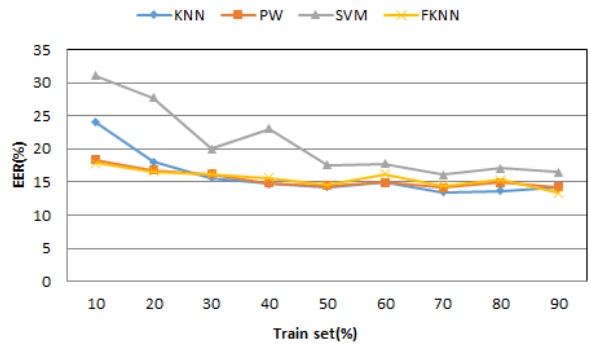

(F)

FIGURE 3. Classification results for different train set divisions (%), Scenario 1 (**A**), Scenario 2 (**B**), Scenario 3 (**C**), Scenario 4 (**D**), Scenario 5 (**E**), and Scenario 6 (**F**) with resultant equal error rates (EER) percentages values (%) for each scenario and classifier employed: K-Nearest Neighbors (KNN), Parzen Window (PW), Support Vector Machine (SVM), and Fuzzy K-Nearest Neighbors (FKNN).

## 3.2. Influence of number of premier features

Performance of classifiers was then assessed based on different numbers of selected premier features. Figure 4 demonstrates equal error rate values based on number of superior features and average of all train sets for 20 iterations. Fuzzy K-Nearest Neighbors demonstrated best overall performance. The results indicate that overall classifier performance was improved when the number of premier features was increased to 15. For higher feature numbers, performance either decreased or did not significantly improve. Using 15 premier features, minimum equal error rates were obtained utilizing the Fuzzy K-Nearest Neighbors classifier in most scenarios ($EER_1$=9.3%, $EER_2$=14.4%, $EER_3$=10.0%, $EER_4$=9.2%, $EER_6$=14.5%).

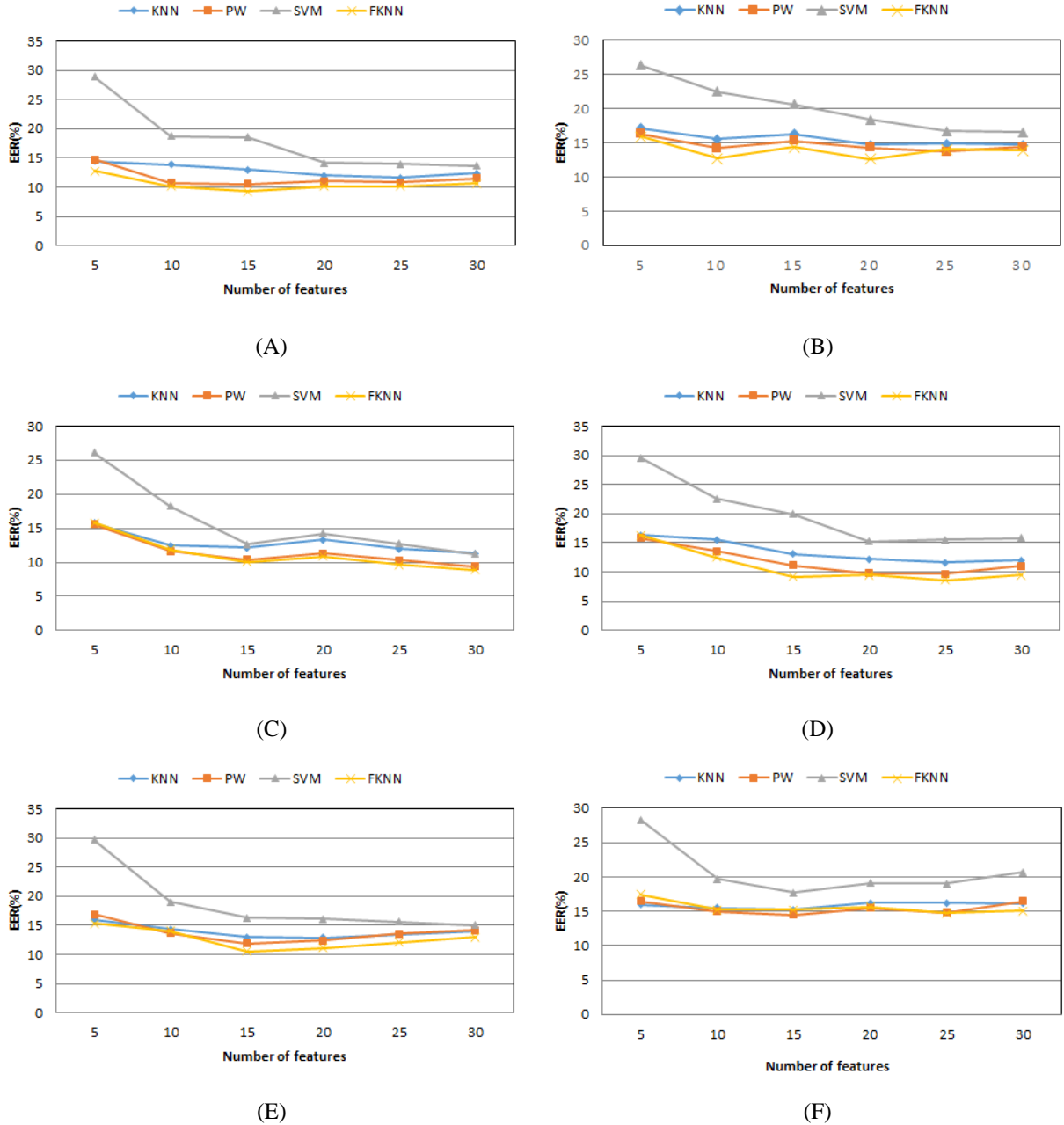

FIGURE 4. Classification results for different number of selected premier features, Scenario 1 (**A**), Scenario 2 (**B**), Scenario 3 (**C**), Scenario 4 (**D**), Scenario 5 (**E**), Scenario 6 (**F**) with resultant equal error rates (EER) percentage values (%) for each scenario and each classifier employed: K-Nearest Neighbors (KNN), Parzen Window (PW), Support Vector Machine (SVM), and Fuzzy K-Nearest Neighbors (FKNN).

### 3.3. Performance at selected working point

Figure 5 provides a comparison of classifier performance for all scenarios when working point is considered as 50% train set size and 15 best features and outlines that Fuzzy K-Nearest Neighbors classifier has the best performance with equal error rates ranging from 8.3%-14.5%. In Table 2 the best 30 features are marked with one asterisk (*) and the best 15 features with two asterisks (**). Six of best 15 features and 12 of the best 30 features involved force while the remaining features were associated with motion.

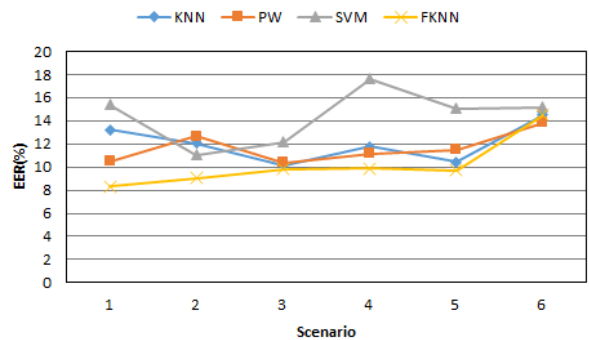

FIGURE 5. Classification results for selected working point (15 best features and 50% train set size) with resultant equal error rates (EER) of average percentage values (%) for each scenario and each classifier employed: K-Nearest Neighbors (KNN), Parzen Window (PW), Support Vector Machine (SVM), and Fuzzy K-Nearest Neighbors (FKNN).

## 4. Discussion

Machine learning is a subset of artificial intelligence, using algorithms (classifiers), which gives computers the capacity to "learn" patterns (progressively improve performance on a specific task) when provided with sufficient data, without needing explicit programming.[26] Supervised, unsupervised, semi-supervised and reinforcement learning algorithms can be used.[26-28] In supervised classifiers, feature data is provided which maximize the ability of classifiers to separate groups by minimizing the error. These techniques have been employed in neurosurgical diagnosis, presurgical planning and outcome prediction.[28] In otolaryngology and dental virtual reality procedures, participants ranged from 1 to 7 skilled (experts) and 5 to 40 novice (less skilled) and differentiated skilled and novice groups from 75 to 100%.

## 4.1. Differentiating skilled and novice performance

Machine learning classifiers had not been used to differentiate skilled and novice groups using virtual reality cerebral tumor procedures. The scenarios utilized in this study involved aspirator skills used in human tumor resections, part of the surgical armamentarium of neurosurgeons and senior residents, but not yet acquired by all junior residents and medical students. It seemed reasonable to define a skilled and novice (less skilled) group based on the required skill set needed for the 6 scenarios studied.[9,25] We applied 4 different supervised machine learning classifiers to the data set involving these participants. Our results demonstrate that all 4 classifier distinguished skilled and novice groups with equal error rates as low as 8.3% indicating the usefulness of classifiers in differentiating participants doing virtual reality procedures. The Fuzzy K-Nearest-Neighbors classifier provided optimal performance and this may relate to its ability to assigns fuzzy rather than crisp membership to the skilled and novice groups.[44] The Support Vector Machine classifier had the least ability to separate groups since it is known to degrade when classifying unbalanced groups.

## 4.2. Misclassification

Table 3 presents the range of individuals misclassified by the Fuzzy K-Nearest-Neighbors classifier. Using this classifier, 19-21 out of 23 skilled individuals and 78-84 out of 92 novices were correctly classified. Some neurosurgeons in this study had cerebrovascular, spinal and functional specialization with little exposure to tumor resection which may be one reason for misclassification. Some junior residents may have been misclassified since they had mastered the required surgical skills. Studies involving more complex scenarios, larger resident numbers and better understanding of which factors and/or combination of metrics to use to better differentiate groups are needed. The potential of machine learning classifiers applied to virtual reality procedures in surgical disciplines is that the new features identified will result in new "metrics" which can then be evaluated in other model systems. These results may not only help us understand the psychomotor skills needed to increase surgical skills but aid in resident assessment and training and improve patient outcomes.

TABLE 3. The range of numbers of individuals correctly and incorrectly classified by the Fuzzy K-Nearest-Neighbors classifier in the 6 different scenarios.

|  | Classified as skilled | Classified as novice | |
|---|---|---|---|
| Skilled | 19-21 | 3-4 | N=23 |
| Novice | 8-14 | 78-84 | N=92 |
|  | N=23 | N=92 | N=115 |

## 4.3. Strengths and Limitations of the Study

The importance of these results lie in their potential educational application to aid in neurosurgical resident training and helping to further define the psychomotor skill sets of expert surgeons.[7] Machine learning and artificial intelligence as applied to virtual reality surgical studies should be seen as useful adjuncts and not a replacement for standard residency training. By relying on 68 features, these machine learning classifiers can automatically capture multiple aspects of psychomotor performance and segregate participants into 'skilled' or 'novice' group. However, this should be seen as an initial step of a formative educational process, prompting instructors to further assess and coach a resident's performance to a desired level.

The classifiers and simulator platform utilized to distinguish neurosurgical skill levels in this study have limitations. First, many of the parametric features included in this investigation have not been assessed in more complex scenarios. Therefore, it is not known if the same classifiers would also be applicable to these scenarios. Whether these parametric features are the most appropriate or other metrics such as the force pyramid or automaticity will be more useful needs to be accessed.[17-20] Second, a simulated aspirator was utilized in the dominant hand which is not representative of the bimanual psychomotor skills and multiple instruments employed during patient tumor resections. Previous studies have demonstrated differences in ergonomics between right and left handed operators and this issue was not addressed in this investigation and deserves further study.[17] Third, the different visual and haptic complexities of simulated tumors utilized and task duration may not adequately discriminate operator performance. More complex and realistic tumor scenarios with simulated bleeding involving use of bimanual instruments are being studied using classifiers which may be more useful. Defining large populations of residents and neurosurgeons with equivalent experience with virtual reality simulation is challenging. Sixteen practicing board certified neurosurgeons from 3 institutions with different areas of expertise participated in this study which is felt to be representative of a general neurosurgical population. We only enrolled residents and medical students from one institution which limits extension of these results. The authors believe that increasing study participants from multiple institutions may further our ability to improve classifier performance to distinguish neurosurgical skill levels at various stages of resident training.

# 5. Conclusion

We presented the first investigation of the application of machine learning in assessing surgical skill level during virtual reality tumor resection. The importance of our results lies in their potential educational application in neurosurgical resident training and helping further define the psychomotor skill set of the skilled surgeon. Machine learning may be one component in helping to realign the present apprenticeship educational paradigm to a more objective model based on proven performance standards.

# References:

1. Kockro RA, Serra L, Tseng-Tsai Y, Chan C, Yih-Yian S, Gim-Guan C, et al. Planning and simulation of neurosurgery in a virtual reality environment. *Neurosurgery*. 2000;46(1):118-37.

2. Bernardo A, Preul MC, Zabramski JM, Spetzler RF. A three-dimensional interactive virtual dissection model to simulate transpetrous surgical avenues. *Neurosurgery*. 2003; 52(3):499–505, discussion 504–505.

3. Radetzky A, Rudolph M Simulating tumour removal in neurosurgery. *Int J Med Inform*. 2001; 64 (2–3):461–472.

4. Lemole GM Jr, Banerjee PP, Luciano C, Neckrysh S, Charbel FT. Virtual reality in neurosurgical education: part-task ventriculostomy simulation with dynamic visual and haptic feedback. *Neurosurgery*. 2007;61(1):142–148; discussion 148–149.

5. Delorme S, Laroche D, DiRaddo R, Del Maestro RF: NeuroTouch: a physics-based virtual simulator for cranial microneurosurgery training. *Neurosurgery*. 2012; 71:32-42.

6. Choudhury N, Gelinas-Phaneuf N, Delorme S, Del Maestro R: Fundamentals of neurosurgery: virtual reality tasks for training and evaluation of technical skills. *World Neurosurg* 2013;80:e9-19.

7. Gelinas-Phaneuf N, Del Maestro RF: Surgical expertise in neurosurgery: integrating theory into practice. *Neurosurgery*. 2013;73 Suppl 1**:**-38.

8. Gelinas-Phaneuf N, Choudhury N, Al-Habib AR, Cabral A, Nadeau E, Mora V, et al: Assessing performance in brain tumor resection using a novel virtual reality simulator. *Int J Comput Assist Radiol Surg.* 2014; 9**:**1-9.

9. Azarnoush H, Alzhrani G, Winkler-Schwartz A, Alotaibi F, Gelinas-Phaneuf N, Pazos V, et al: Neurosurgical virtual reality simulation metrics to assess psychomotor skills during brain tumor resection. *Int J Comput Assist Radiol Surg*. 2015**;**10**:**603-618.

10. Cline BC, Badejo AO, Rivest II, Scanlon JR, Taylor WC, Gerling GJ. Human performance metrics for a virtual reality simulator to train chest tube insertion. *Systems and Information Engineering Design Symposium, SIEDS IEEE*", 2008.

11. Kazemi H, Rappel JK, Poston T, Hai Lim B, Burdet E, Leong Teo C. Assessing suturing techniques using a virtual reality surgical simulator. *Microsurgery.* 2010;30(6):479-486.

12. Trejos AL, Patel RV, Malthaner RA, Schlachta CM. Development of force-based metrics for skills assessment in minimally invasive surgery. *Surg Endosc*. 2014;28(7):2106-19.

13. Kovac ERA, Azhar A, Quirouet J, Delisle, & M. Anidjar. "Construct validity of the lapSim virtual reality laparoscopic simulator within a urology residency program. *Can Urol Assoc J*. 2012;6(4):253-9.

14. Alotaibi FE, Al Zhrani G, Bajunaid K, Winkler-Schwartz A, Azarnoush H, et al. (2015) Assessing Neurosurgical Psychomotor Performance: Role of Virtual Reality Simulators, Current and Future Potential. *SOJ Neurol*. 2015;2(1), 1-7.

15. Alotaibi FE, AlZhrani GA, Mullah MA, Sabbagh AJ, Azarnoush H, Winkler-Schwartz A, et al: Assessing bimanual performance in brain tumor resection with NeuroTouch, a virtual reality simulator. *Neurosurgery*. 11, 2015;Suppl 2:89-98; discussion 98.

16. Alotaibi FE, AlZhrani GA, Sabbagh AJ, Azarnoush H, Winkler-Schwartz A, Del Maestro RF: neurosurgical assessment of metrics including judgment and dexterity using the virtual reality simulator NeuroTouch (NAJD Metrics). *Surg Innov*. 2015;22:636-642.

17. Azarnoush H, Siar S, Sawaya R, et al. The force pyramid: a spatial analysis of force application during virtual reality brain tumor resection. *J Neurosurg*. 2017;127(1):171-181.

18. Sawaya R, Bugdadi A, Azarnoush H, Winkler-Schwartz A, Alotaibi FE, Bajunaid K, AlZhrani GA, Alsideiri G, Sabbagh AJ, Del Maestro RF. Virtual Reality Tumor Resection: The Force Pyramid Approach. *Operative Neurosurgery*. 2017;14(6):686-96.

19. Bugdadi A, Sawaya R, Olwi D, AlZahrani G, Azarnoush H, Sabbagh A, et al: Automaticity of Force Application during Simulated Brain Tumor Resection: Testing the Fitts and Posner Model. *J Surg Educ*. 2018;75(1):104-15.

20. Sawaya R, Alsidieri G, Bugdadi A, Winkler-Schwartz A, Azarnoush A, Bajunaid K, J. Sabbagh AJ, Del Maestro R Development of a Performance Model for Virtual Reality Tumor Resections. *J Neurosurg* [epub ahead of print]. August 3, 2018; DOI: 10.3171/2018.2.JNS172327.

21. Winkler-Schwartz A, Bajunaid K, Mullah MA, Marwa I, Alotaibi FE, Fares J, et al: Bimanual Psychomotor Performance in Neurosurgical Resident Applicants Assessed Using NeuroTouch, a Virtual Reality Simulator. *J Surg Educ*. 2016; 73:942-953.

22. Holloway T, Lorsch Z, Chary M, Sobotka S, Moore MM, Costa AB, et al. Operator experience determines performance in a simulated computer-based brain tumor resection task. *Int J Comput Assist Radiol Surg*. 2015;10(11):1853-1862.

23. Bajunaid K, Mullah MA, Winkler-Schwartz A, Alotaibi FE, Fares J, Baggiani M, et al: Impact of acute stress on psychomotor bimanual performance during a simulated tumor resection task. *J Neurosurg*. 2017;126:71-80.

24. Alzhrani G, Del Maestro RF. *A Validation Study of NeuroTouch in Neurosurgical Training*. Saarbrücken, Germany: Lambert Academic Publishing; 2014.

25. Alzhrani G, Alotaibi F, Azarnoush H, Winkler-Schwartz A, Sabbagh A, Bajunaid K, et al: Proficiency performance benchmarks for removal of simulated brain tumors using a virtual reality simulator NeuroTouch. *J Surg Educ* 2015;72**:**685-696.

26. Samuel AL. Some studies in machine learning using the game of checkers. *IBM Journal of research and development*. 1959;3(3):210-229.

27. Obermeyer Z, Emanuel EJ. Predicting the future - big data, machine learning, and clinical medicine. *N Engl J Med*. 2016;375(13):1216-1219**.**

28. Senders JT, Arnaout O, Karhade AV, Dasenbrock HH, Gormley WB, Broekman ML et al. Natural and artificial intelligence in neurosurgery: a systematic review. *Neurosurgery*. 2017;83(2):181-192.

29. Azimi P, Mohammadi HR, Benzel EC, Shahzadi S, Azhari S, Montazeri A. Artificial neural networks in neurosurgery. *J Neurol Neurosurg Psychiatry*.2015;86(3):251-256.

30. Watson RA. Use of a machine learning algorithm to classify expertise: Analysis of hand motion patterns during a simulated surgical task. *Acad Med*. 2014;89(8):1163-7.

31. Rhienmora P, Haddawy P, Khanal P, Suebnukarn S, Dailey MN. A virtual reality simulator for teaching and evaluating dental procedures. *Methods Inf Med*. 2010;49(4):396-405.

32. Kerwin T, Wiet G, Stredney D, Shen HW. Automatic scoring of virtual mastoidectomies using expert examples. *Int J Comput Assist Radiol Surg*. 2012;7(1):1-11.

33. Ma X, Wijewickrema S, Zhou S, Zhou Y, Mhammedi Z, O'Leary S, et al. Adversarial generation of real-time feedback with neural networks for simulation-based training. arXiv preprint arXiv:1703.01460.

34. Ma X, Wijewickrema S, Zhou Y, Zhou S, O'Leary S, Bailey J. Providing Effective Real-Time Feedback in Simulation-Based Surgical Training. In: Descoteaux M, Maier-Hein L, Franz A,

Jannin P, Collins DL, Duchesne S, eds. *Medical Image Computing and Computer-Assisted Intervention − MICCAI 2017*. Cham: Springer International Publishing, 566-574, 2017.

35. Wijewickrema S, Ma X, Piromchai P, Piromchai P, Briggs R, James Bailey J. et al., Providing Automated Real-Time Technical Feedback for Virtual Reality Based Surgical Training: Is the Simpler the Better? In: Penstein Rosé C, Martínez-Maldonado R, Hoppe HU, et al., eds. *Artificial Intelligence in Education*. Cham: Springer International Publishing; 2018:584-598.

36. Sewell C, Morris D, Blevins NH, Dutta S, Agrawal S, Federico Barbagli F et al. Providing metrics and performance feedback in a surgical simulator, *Comput Aided Surgery*.2008;13:2, 63-81.

37. Rashidi S, Fallah A, Towhidkhah F. Authentication Based on Pole-zero Models of Signature Velocity**.** *J Med Signals Sens*. 2013;3(4):195-208.

38. Rohrer B, Fasoli S, Krebs H, Hughes R, Volpe B, Frontera W et al. Movement smoothness changes during stroke recovery. *J Neurosci.*. 2002;15;22(18):8297-304.

39. Cavallo F, Megali G, Sinigaglia S, Tonet O, Dario P. A biomedical analysis of a surgeon's gesture in a laparoscopic virtual scenario. *Stud Health Technol Inform.* 2006;119:79-84.

40. Trejos Al, Patel RV, Naish MD, Malthaner RA, Schlachta CM. The application of force sensing to skills assessment in minimally invasive surgery. *IEEE International Conference on Robotics and Automation*, 2013.

41. K.Deng. Omega: On-line Memory-based General Purpose System Classifier. The Robotics Institute School of Computer Science Carnegie Mellon University Pittsburgh, Chapter 7, 1998.

42. Ladha L, Deepa T. Feature selection methods and algorithms. *International journal on computer science and engineering.* 2011;3(5):1787-1797.

43. Kung SY. Kernel Methods and Machine Learning. *Cambridge University Press*; 2014, p. 34.

44. Keller JM, Gray MR, Givens JA. A fuzzy k-nearest neighbor algorithm. *IEEE transactions on systems, man, and cybernetics.* 1985(4):580-585.